\documentclass[11pt]{article}

\usepackage[final]{acl}

\usepackage{times}
\usepackage{latexsym}

\usepackage[T1]{fontenc}

\usepackage[utf8]{inputenc}

\usepackage{microtype}

\usepackage{inconsolata}

\usepackage{hyperref}
\usepackage{url}
\usepackage{graphicx} 
\usepackage[hypcap=false]{caption}
\usepackage{booktabs}
\usepackage[table]{xcolor}
\usepackage{wrapfig}
\usepackage{listings}
\usepackage{booktabs}
\usepackage{multicol}
\usepackage{multirow}
\usepackage{colortbl}
\usepackage{enumitem}
\usepackage{amsmath}

\usepackage{amsfonts}
\usepackage{amssymb}
\usepackage{amsmath}   
\usepackage{amsthm}    

\usepackage{makecell} 
\usepackage{bm} 
\usepackage{amssymb} 

\newcommand{\E}{\mathbb{E}}
\newcommand{\sR}{\mathbb{R}}
\newcommand{\vone}{\mathbf{1}}

\usepackage{makecell} 
\usepackage{bm} 
\usepackage{amssymb} 
\usepackage{amsmath} 

\definecolor{LightRed}{HTML}{F4CCCC}
\definecolor{LightYellow}{HTML}{FFF2CC}
\definecolor{LightGreen}{HTML}{D9EAD3}

\title{Beyond Cross-Modal Alignment: Measuring and Leveraging Modality Gap in Vision-Language Models}

\author{
 \textbf{Hanqi Yan\textsuperscript{1,*}},
 \textbf{Xiangxiang Cui\textsuperscript{2,*}},
 \textbf{Lu Yin\textsuperscript{2}},
 \textbf{Jindong Gu\textsuperscript{3}},
\\
 \textbf{Paul Pu Liang\textsuperscript{4}},
 \textbf{Yulan He\textsuperscript{1,5,\textdagger}},
 \textbf{Yifei Wang\textsuperscript{4,\textdagger}}
\\
\\
 \textsuperscript{1}King's College London, UK,
 \textsuperscript{2}University of Surrey, UK,
 \textsuperscript{3}University of Oxford, UK,
\\
 \textsuperscript{4}MIT CSAIL, USA,
 \textsuperscript{5}The Alan Turing Institute, UK
\\
 \small{
 \textbf{Correspondence:} \href{mailto:yulan.he@kcl.ac.uk}{yulan.he@kcl.ac.uk}, \href{mailto:yifei_w@mit.edu}{yifei\_w@mit.edu}
 }
}

\begin{document}
\maketitle
\renewcommand{\thefootnote}{}
\footnotetext{*Equal contributions. \textdagger Corresponding authors.}
\renewcommand{\thefootnote}{\arabic{footnote}}
\begin{abstract}
The success of vision-language models is primarily attributed to effective alignment across modalities such as vision and language. However, modality gaps persist in existing alignment algorithms and appear necessary for human perception -- evident in modality-specific phenomena like visual texture and linguistic tone. These observations motivate us to computationally measure and leverage modality gaps to improve downstream tasks. We first introduce the \textbf{M}odality \textbf{D}ominance \textbf{S}core (\textbf{MDS}), which attributes multimodal features to specific modalities by categorizing them into three classes: vision-dominant features, language-dominant features, and cross-modal features. We then propose automatic interpretability metrics to evaluate these modality-specific features in a scalable manner. Finally, we demonstrate that the training-free model editing enhances multiple downstream tasks, including mitigating bias in gender classification, generating cross-modal adversarial examples, and enabling modality-specific control in text-to-image generation. Combined with task-agnostic interpretability tools, our work offers insights for systematic analysis and lightweight editing of multimodal models.
\end{abstract}
\section{Introduction}
\label{sec:intro}
Multimodal models have become foundational to the advancement of AI, enabling AI systems to process and understand information from multiple data modalities, such as vision and language~\cite{radford2021learning, kim2021vilt, lu2019vilbert,liang2024foundations}. Vision-Language Models (VLMs) in particular operate under the premise that different data modalities share common, or cross-modal, features that can be jointly learned~\citep{ngiam2011multimodal,sun2024aligning,li2025unifiedmllm}. 

Alongside these remarkable advancements, ongoing AI research aims to deepen our understanding of how different modalities interact and diverge within VLMs~\citep{liang2022mind,Rawal2023DissectingMI,schrodi2025two,zhang2025evaluating}. For instance,\citet{liang2022mind} revealed that image and text embeddings often reside in disjoint regions of the shared embedding space. \citet{parcalabescu2023mm} proposed Shapley value-based attribution methods to quantify the extent to which multimodal models rely on individual modalities, with follow-up work~\citep{parcalabescu2025do} diagnosing unimodal collapse. 

Despite these modality preference measurements and optimization methods, existing studies treat modality gaps as undesirable imperfections, primarily diagnosing model collapse~\citep{parcalabescu2025do,Rawal2023DissectingMI,schrodi2025two} or motivating new training algorithms for improved alignment.
Our work takes a contrasting perspective: we posit that \textit{modality gaps} are both prevalent and beneficial for downstream tasks. This assumption is grounded in cognitive science, where modality commonality and separation have long been central themes. Researchers have examined how humans integrate and differentiate information across sensory modalities~\citep{paivio1991dual,spence2011crossmodal,article}, suggesting that modal specificity may be functionally advantageous rather than merely an artifact of imperfect alignment.\\
To investigate this hypothesis in VLMs, we conduct a systematic study with three core contributions:
\begin{itemize}[topsep=-2pt, leftmargin=6pt,itemsep=-2pt]
\item[] 1. We demonstrate that modality-specific information can be extracted from VLMs—specifically, text-dominant (\texttt{TextD}), image-dominant (\texttt{ImgD}), and cross-modal (\texttt{CrossD}) features—and show that these features exhibit distinct activation patterns when processing images versus text.
\item[] 2. We propose embedding-based interpretability metrics to measure monosemanticity (within-modality coherence) and modality fidelity (cross-modality validation) in a multimodal setup. These metrics are scalable and compatible with the existing top-k activated interpretations.
\item[] 3. We design lightweight probing and steering methods to analyze models' concept-specific preferences and achieve precise, effective control over VLM behavior.
\end{itemize}

\section{Related Work}
\paragraph{Modality Gap.} The study of modality differences has long been explored in cognitive science~\citep{spence2011crossmodal, paivio1991dual, calvert2004handbook}. In multimodal models, researchers have identified that modality bias and gaps are prevalent in both early Multimodal Models (MMs)~\citep{liang2022mind} and large MMs~\citep{zhang2025evaluating}. Moreover, modality gaps have been shown to negatively impact downstream tasks such as video understanding~\citep{Rawal2023DissectingMI} and object detection~\citep{schrodi2025two}. Several metrics have been proposed to quantify modality gaps, including L2M~\citep{liang2022mind}, MM-SHAP~\citep{parcalabescu2023mm}, and its variants~\citep{parcalabescu2025do}, which measure the degree to which individual modalities contribute to model predictions. Our work differs in two key aspects: (i) we measure modality gaps at the component level, identifying how individual model components respond differentially to different modalities; (ii) we leverage modality-specific components for probing and steering, treating specialization as a functional feature rather than an imperfection.\\
\paragraph{Interpretability Measurements.}
Existing interpretability measurements are based on summarizing patterns in top-k activated samples. For example, logit lens~\citep{logitlens} has inspired many studies in both unimodal and multimodal representation understanding~\citep{parekh2024concept,jiang2025interpreting}. The embedding-based extension~\citep{phukan-etal-2025-beyond} alleviates its limitation in processing contextual-related concepts. More recently, LLMs have been used to generate explanations for activation patterns, with prediction accuracy on held-out samples serving as an interpretability proxy~\citep{bills2023language}. However, this LLM-as-a-judge approach is computationally expensive and only measures semantic coherence within the unimodality, neglecting cross-modality consistency. Our interpretability metrics address these limitations through scalable embedding-based computation and introduce modality fidelity to fill the gap.

\section{Identify Modality-Specific Features}
\label{sec:mds}
Modality alignment and fusion are crucial to the success of existing VLMs~\citep{liang2022mind,schrodi2025two}, while the modality-specific gap has been extensively studied in cognitive science. For instance, \citet{UNGERLEIDER1994157,article} have found that regional specificity and coordinated processing coexist in the human brain. Therefore, we start with the question \textit{``whether there are modality-specific features in VLMs?''} To answer the question, we use CLIP models from OpenAI~\citep{radford2021learning}.

\paragraph{Background: Modality Alignment in VLMs.}  Typically, there are an image encoder and a text encoder in a VLM for image and text input processing, respectively. Specifically, the image-text pair $(x_\text{img}, x_\text{txt})$ is fed to an image encoder $f_\text{img}$ and a text encoder $f_\text{txt}$ within the model, respectively and the final-layer representations $z_\text{img}\in\mathbb{R}^{D}$ and $z_\text{txt}
\in\mathbb{R}^{D}$ are then optimized jointly in the shared $D$-dimensional representation space. An alignment loss, such as the contrastive loss in CLIP~\citep{ilharco_gabriel_2021_5143773} across the two modalities, is applied for modality alignment. 
A persistent modality preference/gap remains across most multimodal models~\citep{liang2022mind,zhang2025evaluating}. 

\subsection{Modality-specific Feature Identification}
\label{subsec:MDS}

To measure the modality gap, \citet{liang2022mind} used the difference between the center of image embeddings and text embeddings of $M$ input pairs, i.e., $\frac{1}{M}(\sum_{i=1}^{M} ||z_{\text{img}, i}||_{2}-\sum_{i=1}^{M} ||z_{\text{txt},i}||_{2})$. We extend this model-level measurement to a fine-grained metric, i.e., the predominant modality associated with each dimension $d\in \{1,2,\dots,D\}$ in the shared embedding space. The proposed modality dominance score (\textbf{MDS}), denoted as $R(d)$ shown in Eq.~(\ref{eq:mds}) reflects how strongly the $d$-th feature~\footnote{Each feature dimension corresponds directly to a feature/neuron in the VLM’s final layer; our study thus focuses on the interpretability of the model’s intrinsic components.} is influenced by the image modality:
\vspace{-1ex}
\begin{align}
\scriptsize
   R(d)=\frac{1}{M}\sum_{i=1}^M\frac{||z^{(d)}_{\text{img},i}||}{|| z^{(d)}_{\text{img},i}||+||z^{(d)}_{\text{txt}, i}||}. 
\label{eq:mds}
\end{align}
Specifically, we feed $M$ image-text pairs to the VLM and extract the corresponding image features $z_{\text{img},i}$ and text features $z_{\text{txt},i}$ for $i$-th input. For each $d$-th dimension in the $D$-dimension shared space, we calculate the relative activation between the features from the two modalities. This modality fraction is averaged over more than $M=10k$ input pairs, providing a representative estimate of the modality distribution.\footnote{Details of the MDS calculation are in Appendix~\ref{app:mds}.}

We then categorize all $D$ features into three groups based on their deviation from the mean $\mu$ and standard deviation $\sigma$ of the MDS distribution: 
\begin{equation}
\begin{aligned}
\texttt{TextD: } &R(d) < \mu-\sigma; \\
\texttt{CrossD: } &\mu-\sigma < R(d) < \mu+\sigma; \\
\texttt{ImgD: } &R(d) > \mu+\sigma
\end{aligned}
\end{equation}
We anticipate that \texttt{ImgD} features are predominantly activated by visual concepts, \texttt{TextD} features by textual concepts, and \texttt{CrossD} features are simultaneously activated by the shared commonalities between image and text. 

\subsection{Quantitative Evaluation for MDS}
To verify that modality-specific features effectively capture their intended modality information, we employ an ablation-based validation approach. Specifically, we remove these features from the original representation from CLIP ViT-H/14 (LAION-2B)~\cite{ilharco_gabriel_2021_5143773} by zeroing out their corresponding indices, then use the modified representations as input features for logistic regression.  We evaluate the representation on image/text classifications using samples from COCO~\citep{lin2014microsoft}. A decrease in classification accuracy indicates that the removed features contained substantial modality-specific information, thereby validating our feature attribution method.

The results are shown in Table~\ref{tab:ablation}. It is observed that removing the \texttt{ImgD} leads to larger classification degradation in Image classification, while removing \texttt{TextD} leads to larger drops in text classification; while \texttt{CrossD} does not show any particular modality tendency in classification.
\renewcommand{\arraystretch}{0.6}
\begin{table}[ht]
\centering
\tiny
\resizebox{0.45\textwidth}{!}{%
\begin{tabular}{p{0.3cm}lccr}
\toprule
\textbf{Task} & \textbf{Deletion} & \textbf{\# Neurons} & \textbf{Accuracy} & \textbf{$\Delta$ Acc} \\
\midrule
\multirow{7}{*}{\makecell{Image \\ CLS}} 
 & None & 0 & \textbf{0.776} & / \\
\cmidrule{2-5}
 & Random & 426 & 0.757 & -2.1\% \\
 & \texttt{ImgD} & 426 & \textbf{0.750} & \textcolor{black}{\textbf{-2.6\%}} \\
\cmidrule{2-5}
 & Random & 554 & 0.756 & -2.0\% \\
& \texttt{TextD} & 554 & 0.760 & -1.0\% \\
\cmidrule{2-5}
 & Random & 44 & 0.773 & -0.3\% \\
& \texttt{CrossD} & 44 & 0.769 & -0.5\% \\
\midrule
\multirow{7}{*}{\makecell{Text \\ CLS}}
 & None & 0 & \textbf{0.713} & / \\
\cmidrule{2-5}
& Random & 426 & 0.694 & -1.9\% \\
& \texttt{ImgD} & 426 & 0.702 & -1.1\% \\
\cmidrule{2-5}
& Random & 554 & 0.693 & -2.0\% \\
& \texttt{TextD} & 554 & \textbf{0.683} & \textcolor{black}{\textbf{-3.0\%}} \\
\cmidrule{2-5}
& Random & 44 & 0.710 & -0.3\% \\
& \texttt{CrossD} & 44 & \textbf{0.712} & -0.1\% \\
\bottomrule
\end{tabular}
}
\caption{Performance changes of Modality-specific classification after removing: random vs. specialized features, i.e., \texttt{ImgD}, \texttt{TextD} and \texttt{CrossD}. We also remove the same number of random feature indices for comparison.}
\label{tab:ablation}
\end{table}

\begin{figure*}[h]
\centering
\includegraphics[width=0.65\linewidth,trim={0 0 120 0},clip]{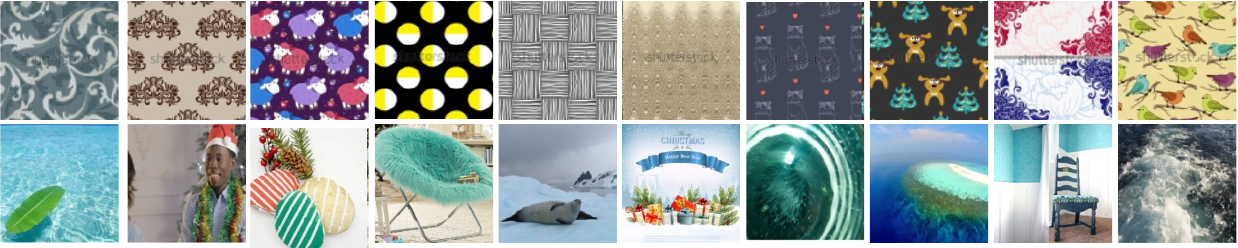}
\label{fig:act_imgs_imgD}
\vfill
\resizebox{0.65\textwidth}{!}{%
\begin{tabular}{p{7.5cm}|p{7.5cm}}
\toprule[1.pt]
\small
{\textit{\textbf{Feature-647: Pattern and others.}}} & {\textit{\textbf{Feature-667: Scenes in winter and other.}}} \\
\toprule[1.pt]
\textcolor{blue}{Seamless pattern, flowers on a background.}&Covering the trailhead in a \textcolor{blue}{winter} wonderland. \\
\midrule
Every girl should have this in their bedroom.&Red leather belt, a perfect accessory. \\
\midrule
Could new showroom and model signal the start?&The image of drum under the white background.\\
\bottomrule
    \end{tabular}
    }
\vspace{-3mm}
\captionof{figure}{ Activated images and texts (in Table) by \texttt{ImgD}. Top image row (feature 647): patterns and textures. Bottom image (feature 667): water and aquatic themes in blue. Texts in \textcolor{blue}{blue} align with visual concepts.}
\label{tab:act_txts_imgD}
\end{figure*}
\begin{figure*}[h]
\centering
\includegraphics[width=0.65\linewidth,trim={0 0 120 0},clip]{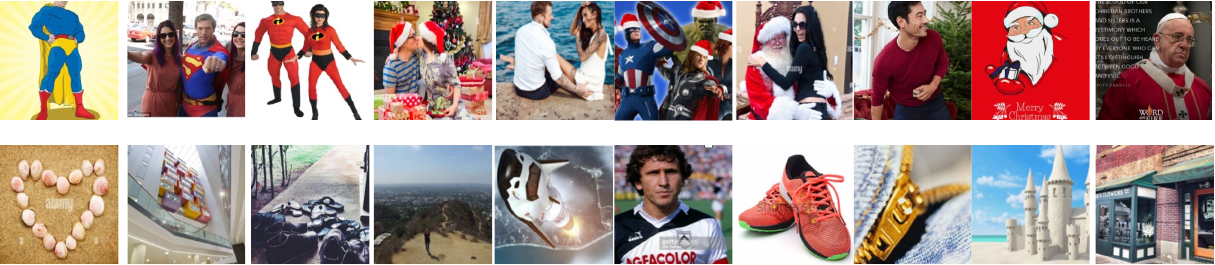}
\label{fig:act_imgs_textD}
\vfill
\vspace{-0mm}
\resizebox{0.65\textwidth}{!}{%
\begin{tabular}{p{6.8cm}|p{7.3cm}}
\toprule
\textit{\textbf{Feature-34: Sweet and happy Couple.}} & \textit{\textbf{Feature-242: Strong emotion.}} \\
\midrule
Attractive young couple sitting on a bench, talking and \textcolor{blue}{laughing} with the city. & Animal looking for a cat tree without carpet your options have \textcolor{blue}{greatly} expanded. \\
\midrule
Sculpture of \textcolor{blue}{lovers} at the temple &Sinkhole, \textcolor{blue}{most terrifying thing I have ever seen.}\\
\midrule
Young couple in \textcolor{blue}{love}, \textcolor{blue}{hugging} in the old part of town.&We're away from the beginning of the \textcolor{blue}{holiday season} here\textcolor{blue}{!}\\
\bottomrule
    \end{tabular}
    }
\vspace{-2mm}
\captionof{figure}{ Activated images and texts (in Table) by \texttt{TextD}. Top image row (feature 34): couples and individuals in red attire. Bottom image row (feature 242): diverse objects. Text in \textcolor{blue}{blue} aligns with visual concepts.}
    \label{tab:act_txts_TextD}
\end{figure*}
\label{subsec:vis_acts}
\subsection{Qualitative Evaluation for MDS}
We randomly select two features from the three groups, and then display their most-activated images and texts in Figure~\ref{tab:act_txts_imgD} (\texttt{ImD}), Figure~\ref{tab:act_txts_TextD} (\texttt{TextD}) and Figure~\ref{tab:act_texts_crossm} (\texttt{CrossD}).
\textbf{\texttt{ImgD} activates fundamental visual concepts, such as repeated patterns and colors.}
Feature 647 activates images with diverse repetitive patterns; feature 667 focuses on scenes with aquatic-blue elements. Although less coherent than the images, some patterns do emerge for its activated texts: feature-647 activates two sentences that refer to repetitive patterns, such as ``\emph{tufted upholstery}''; feature-667 activates texts related to ``\emph{snowy}'' and ``\emph{winter}''. These observations indicate the modality alignment while the visual commanalities are more predominant for the \texttt{ImgD}. 
\textbf{\texttt{TextD} capture abstract concepts, such as human feelings and atmosphere.} For the activated images for feature-34 (the 1st row), most of the images have red color, with one image depicting a couple talking beside the sea; for feature-242, there are no clear patterns among the activated images. When looking at the activated texts, sentences activated by feature-34 center around a sweet and happy atmosphere between couples, with themes like cuddling, embracing, and hugging. Feature-242 focuses on strong human emotions, such as ``\textit{never}'', ``\textit{terrifying}'' and exclamation marks. These \texttt{TextD} generally correspond to abstract and consistent human emotions, which can be conveyed with a variety of visual objects. For example, in the second row, the first image depicts a collection of stones forming a heart shape, while the fourth image is a scenic view during a great trip. 
\textbf{\texttt{CrossD} (the majority features) capture shared semantics across modalities.} Differently, \texttt{CrossD} features capture common concepts that could be expressed in both visual and language modalities. (Details can be found in Appendix~\ref{app:crossD_qualitative}.)

\section{Automatic Interpretability Evaluation}
\label{sec:mono}
Although we have identified modality-dominant features, features in deep models are inherently \emph{polysemantic}~\citep{olah2020zoom}—each feature often encodes multiple unrelated semantic concepts, potentially spanning both textual and visual modalities, hindering interpretability. \emph{Monosemanticity}~\citep{elhage2022solu,bills2023language,Gurnee2023FindingNI,Yan2024EncourageOI} has thus emerged as a paradigm for deriving interpretable features that encode single concepts. However, scaling interpretability evaluation remains an open challenge due to heavy reliance on costly human annotations~\citep{gao2024scaling} or LLM-generated explanations~\citep{bills2023language}. To address this bottleneck, we propose a suite of automated metrics to measure feature interpretability in multimodal models.

\subsection{Overview}

A feature is considered interpretable if its semantic meaning can be readily understood by humans. In practice, interpretability is assessed by examining whether the top-k activated samples exhibit coherent and consistent patterns. This top-k activation-based interpretation has become standard practice for analyzing both language~\citep{geva-etal-2021-transformer} and multimodal models~\citep{parekh2024concept}. \citet{bills2023language} advanced this approach by first prompting a large language model (LLM) to generate explanations based on activated tokens, then using these explanations to predict activation values for held-out tokens. The correlation between predicted and actual activations serves as an interpretability score. However, this LLM-as-a-judge paradigm faces limitations in reliability and scalability~\citep{gu2024survey,yan-etal-2024-mirror}.

Building upon the top-k activation framework, we propose scalable, \textit{embedding-based} evaluation metrics tailored to multimodal models. In this context, interpretability encompasses two dimensions:
\begin{itemize}[itemsep=0pt,topsep=0pt,leftmargin=0pt]
\item \emph{\textbf{Monosemanticity}} (within-modality coherence). It measures whether a feature's top-k activated samples exhibit semantic coherence within a single modality based on their embedding similarity.
\item \emph{\textbf{Modality fidelity}} (cross-modality validation). It compares the monosemanticity scores across modality-attributed features to assess whether features remain faithful to their assigned modality. For example, when processing visual inputs, \texttt{ImgD} should exhibit higher monosemanticity scores than \texttt{TextD}, and vice versa for textual inputs.
\end{itemize}


\subsection{Intepretability Metrics}

\label{subsec:mono_eval}
Given a feature $z^{(d)}$, the $d$-th dimension of $z\in\mathbb{R}^{D}$, we propose to use embedding models $h: \mathbb{Z}^{D} \rightarrow \mathbb{Z}^{D'}$ to calculate the interpretability metrics.

\paragraph{Monosemanticity.} 
For each image/text feature $z^{(d)}$, we collect the top $m$ most-activated image/text samples for this dimension, and feed them to the embedding model $h$ to get $Z_+\in\sR^{m\times D'}$. For comparison, we embed $m$ random samples into $Z_-\in\sR^{m\times d'}$. Then, we calculate the inter-sample similarity between the selected samples, $S_+=Z_+Z_+^\top\in\sR^{m\times m}$ and $S_-=Z_-Z_-^\top\in\sR^{m\times m}$. The monosemanticity of an individual feature $z^{(d)}$ is measured by calculating the relative difference between the two similarity scores, denoted as $I(z^{(d)})$ (\textit{\textbf{EmbSim}}). We also propose a binary metric to avoid the different scales in different modalities, denoted as $W(z^{(d)})$ (\textbf{\textit{WinRate}}):
\begin{equation}
\scalebox{0.85}{$\begin{aligned}
I(z^{(d)}) &= \frac{1}{m(m-1)}\sum_{i\neq j}\frac{(S_+)_{ij}-(S_-)_{ij}}{(S_-)_{ij}}, \\
W(z^{(d)}) &= \frac{1}{m(m-1)}\sum_{i\neq j}\vone_{[(S_+)_{ij}>(S_-)_{ij}]}.
\end{aligned}$}
\end{equation}
The overall interpretability score is the average across all $d$ dimensions for $z^{(d)}$ where $d \in [1, D]$. A higher monosemanticity score (both \textit{EmbSim} and {\textit{WinRate}} are \textbf{\textit{Mono}}, with a superscript representing the input modality) indicates that the extracted features exhibit stronger semantic consistency towards the given modality. \footnote{We calculate the average of \textit{EmbSim} and \textit{WinRate} as \textit{Mono} in the main content; the separate results for the two metrics can be found in Appendix~\ref{app:mono_results}.}

\paragraph{Modality Fidelity.} Based on the single-modality monosemanticity (semantic coherence), we proceed to cross-modality interpretability. Specifically, we ask: {is \texttt{ImgD} indeed more effective at capturing coherent visual inputs than \texttt{TextD}?} Similarly, is \texttt{TextD} better at encoding textual semantics compared to \texttt{ImgD}?  Therefore, we define the modality fidelity as:
\begin{equation}
\scalebox{0.95}{$\begin{aligned}
     \textit{Visual Fidelity} &=  \textit{Mono}^{\mathrm{vis}}(\texttt{ImgD})-\textit{Mono}^{\mathrm{vis}}(\texttt{TextD}) \\ 
     \textit{Textual Fidelity} &= \textit{Mono}^{\mathrm{txt}}(\texttt{TextD})-\textit{Mono}^{\mathrm{txt}}(\texttt{ImgD})
\end{aligned}$}
\nonumber
\end{equation}


\subsection{Interpretability Evaluation Results}
\label{subsec:mono}

Sparse Autoencoders (SAEs)~\citep{cunningham2023sparse} have been shown to effectively generate monosemantic features by enforcing feature sparsity. Therefore, we apply our metrics to compare CLIP and CLIP+SAE to validate whether our evaluation framework yields conclusions consistent with existing literature, establishing its reliability as an interpretability metric. Beyond this validation, we also explore whether other representation learning methods can enhance feature interpretability.
\subsubsection{Comparison Models}
We incorporate several representation learning algorithms (including SAEs) that aim to learn modality-specific features~\footnote{Implementation details for these methods are shown in Appendix~\ref{app:extract_mono}.}.\\
\noindent\textbf{Multimodal SAEs.} SAEs have emerged as a scalable tool for transforming polysemantic \emph{neurons} into interpretable, monosemantic \emph{features} across various LLMs~\citep{templeton2024scaling,gao2024scaling,lieberum2024gemma}. We extend it for multimodal settings by training a \textbf{single} SAE model $g:\mathbb{Z}\to\mathbb{Z}$ to reconstruct $z$, i.e., the final-layer outputs from the image and text encoder within CLIP, respectively. Specifically, we adopt that applies a linear encoder $W_\text{enc}$ followed by a $\operatorname{Top}K$ operation that only keeps the $K$ most activated units while zeroing out the rest. The sparse latent representation $z^\text{sae}$ is then reconstructed using a linear decoder $W_{\mathrm{dec}}$:
{\abovedisplayskip=4pt \belowdisplayskip=4pt
\begin{equation}
\begin{aligned}
z^\text{sae} &= \operatorname{TopK}\left(W_{\text{enc}}\left(z-b_{\text{pre}}\right)\right), \\
\hat{z} &= W_{\mathrm{dec}} z^\text{sae}+b_{\mathrm{pre}}.
\end{aligned}
\end{equation}}%
$z\in\mathbb{R}^{D}$ is the inputs of SAE, i.e., $z_\text{img}$ or $z_\text{txt}$. $z^\text{sae}\in\mathbb{R}^{n}$ is the learned sparse representation. We train the multimodal SAE to reconstruct $z_i$, $z_t$.\\
\noindent\textbf{DeCLIP.} Beyond multimodal supervision (image-text pairs), DeCLIP~\citep{li2022supervision} also incorporates single-modal self-supervision (image-image pairs and text-text) for more efficient joint learning. \\
\noindent\textbf{Multimodal NCL.} 
As shown in \citet{wang2024ncl}, the non-negative constraints allow Non-negative Contrastive Learning (NCL) to extract highly sparse features and significantly improve feature monosemanticity. 
Therefore, we introduce a variant of NCL to enhance modality specification with the following loss,
\begin{align}
    -\E_{z_\text{img},z_\text{txt}}\log\frac{\exp(g(z_\text{img})^\top g(z_\text{txt}))}{E_{z^-_\text{txt}}\exp(g(z_\text{img})^\top g(z^-_\text{txt}))},
\label{eq:ncl_loss}
\end{align}
where here we use a ReLU-activated MLP network $g$ to map input features to \emph{non-negative} output latent features.

\begin{figure*}[t]
  \centering
  \includegraphics[width=0.85\linewidth,trim={130 120 140 130},clip,]{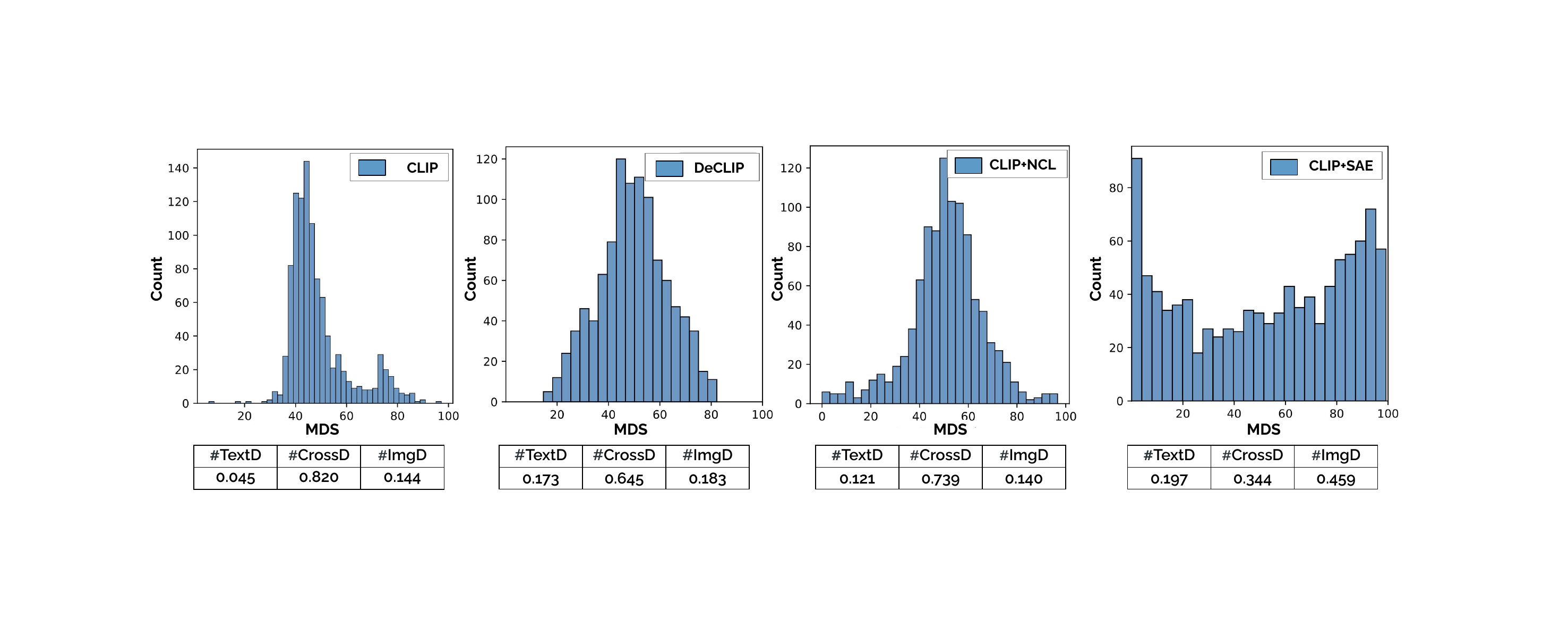}
  \caption{Modality Dominance Score (MDS) distributions of three feature categories for different VLMs.}
  \label{fig:mds}
\end{figure*}

\subsubsection{Evaluation Results}
\begin{figure}[htbp]
\centering
\begin{minipage}{0.45\textwidth}
\centering
\includegraphics[width=0.99\textwidth,trim={0 23 0 22},clip]{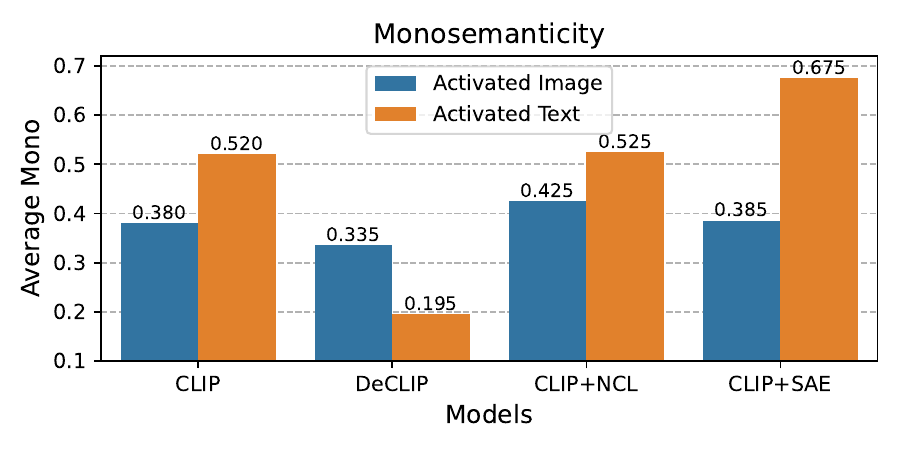}
\vspace{-8mm}
\caption{\footnotesize Monosemanticity.}
\label{fig:interpret_metrics_allmodels}
\end{minipage}
\hfill
\begin{minipage}{0.45\textwidth}
\centering
\includegraphics[width=0.99\textwidth,trim={0 23 0 22},clip]{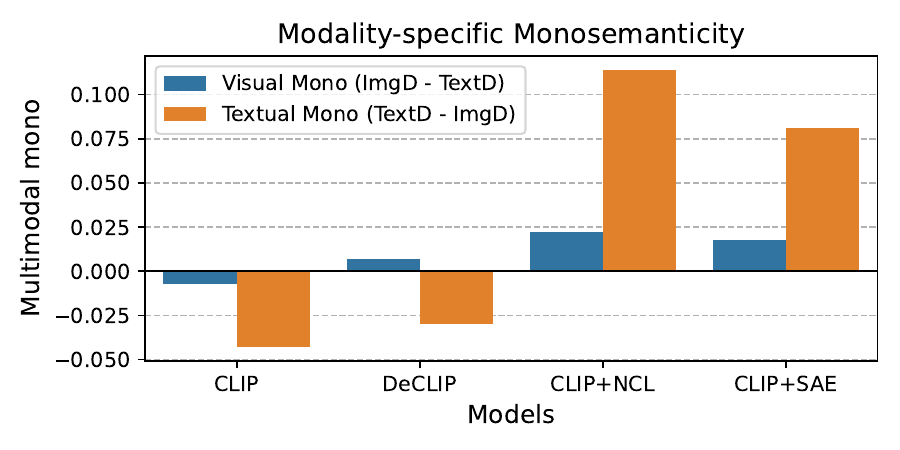}
\vspace{-8mm}
\caption{\footnotesize Modality Fidelity.}
\label{fig:relative_interpret}
\end{minipage}
\vspace{-4mm}
\end{figure}

We calculate the monosemanticity and modality fidelity for all the models. \\
\noindent\textbf{Results of Monosemanticity.} We compute the {\textit{Mono}} (interpretability) score by identifying the top-20 most activated images and texts for each feature, respectively. From the average interpretability results in Figure~\ref{fig:interpret_metrics_allmodels}, we observe the following: (i) The features extracted using SAE and NCL (which enforce the feature sparsity) exhibit the highest overall monosemanticity for both activated input images and texts. (ii) DeCLIP does not enhance interpretability through self-supervision alone; the monosemanticity on the textual side becomes even worse. This suggests that polysemantic features remain prevalent in DeCLIP, although their modality separation is clearer than in CLIP.\\
\noindent Moreover, we observe that \textbf{monosemanticity enhancement encourages more modality-specific neurons}. In Figure~\ref{fig:mds}, we calculate the MDS and visualize the distributions of the three feature groups across models. Interestingly, we find that CLIP contains a spectrum of features with different modality dominance. Specifically, its distribution skews towards the image modality, and this trend is consistent across all models. DeCLIP, on the other hand, shows a more balanced and less centered distribution. This suggests that DeCLIP, through self-supervision, extracts more modality-specific features, which might be overlooked by pure vision-language contrastive models like CLIP. 
The extracted features on top of NCL and SAE also exhibit less skewness, with SAE showing the most balanced distribution, indicating its strong capability to extract diverse monosemantic features. \\
\noindent\textbf{Results of Modality Fidelity.} We have the following observations from Figure~\ref{fig:relative_interpret}: 
(i) For CLIP,  all the modality monosemanticity is negative, demonstrating the high entanglement of the two modality information. (ii) All the methods prompt the modality monosemanticity compared with CLIP. Particularly, the improvements of DeCLIP can be attributed to its single-modal alignment training loss, which could weaken some cross-modal associations in CLIP. (iii) NCL stands out as the best model for capturing both visual and textual monosemantic features, followed by SAE.

\section{Applications of Modality Gap}
\label{sec:case_study}
Beyond interpretability, we design lightweight probing and steering methods based on modality-specific features to analyze VLMs' perceptual biases and enable precise behavioral control.~\footnote{Detailed implementations are in Appendix~\ref{app:case_study}.} 


\subsection{Understanding Gender Pattern}
\label{subsec:gender}
We describe gender using both visual and textual features and these data are used to train VLMs. To test whether there is a modality-specification in different genders, e.g., \textit{Does the concept of the feminine get described by images more frequently? such as more colorful outfits.} To test this hypothesis, we collect both male and female images with their corresponding textual descriptions from the cc3m-wds~\citep{sharma2018conceptual}. These images are then encoded using the Clip+SAE model, extracting 1024-dimensional features for both female and male subjects. Next, we apply a zero-mask intervene strategy to remove the \texttt{ImgD} and \texttt{TextD} from these representations. 

We compare changes in gender classification accuracy when removing \texttt{ImgD} features from image inputs, which capture dominant feminine visual cues, versus removing \texttt{TextD} features from text inputs. As shown in Table~\ref{tab:gender_modality}, we find that feminine concepts are primarily preserved in \texttt{ImgD} (as the removal of \texttt{ImgD} from the image leads to larger classification degradation), whereas male concepts are more affected by the removal of \texttt{TextD}. 
\begin{table}[h]
    \centering
    
    \resizebox{0.30\textwidth}{!}{%
    \begin{tabular}{r|rr}
    \toprule
    \textbf{Gender}    & w.o ImgD & w.o TextD \\
    \midrule
    Female       & \textbf{17.65}&7.27\\
    Male & 5.64&\textbf{28.67}\\
    \bottomrule
    \end{tabular}
    }
    \caption{Gender classification changes (\%) after removing \texttt{ImgD}(\texttt{textD}) \textbf{from input image(text)} for both female and male concepts identification. It is to verify the dominant modality for different genders.
    }
    \label{tab:gender_modality}
\end{table}
\begin{figure}[ht]
    \centering
    \includegraphics[width=0.48\textwidth,trim={70 400 80 50},clip]{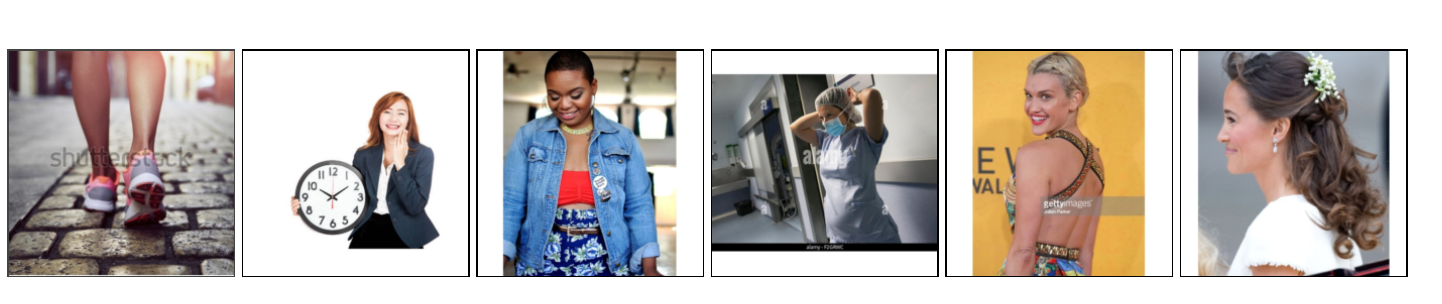}
    \caption{\textit{\textbf{Female}} figures ordered by their percentages of \texttt{ImgD} features: 0.14, 0.16, 0.18, 0.20, 0.22, 0.24, 0.26. More feminine concepts are observed to be related with more \texttt{ImgD}.}
    \label{fig:female_diff_imgd}
\end{figure}

\noindent\textbf{Understand the what feminine concepts the \texttt{ImgD} represent.} We sample different female images which differ in the percentage of their most activated features categorized as \texttt{ImgD} features. The results are in Figure~\ref{fig:female_diff_imgd}. From left to right, more activated features are \texttt{ImgD}, and they tend to contain more detailed (\textit{stereotype}) feminine concepts, such as a backless skirt, hair accessories. The middle images show professional females, such as a politician and a doctor; and the first image shows a pair of legs in sports shoes, with minimal feminine factors, the pink color. 

\subsection{Generate Modality-Specific Attacks}
\label{subsec:ad_att}
We investigate the impact of different types of features on multimodal adversarial attacks~\citep{cui2024robustness,yin2024vlattack}, following the setup in~\citet{shayegani2024jailbreak}.

\begin{figure}[h]
    \centering
    \includegraphics[width=0.95\linewidth,trim={20 12 20 5},clip]{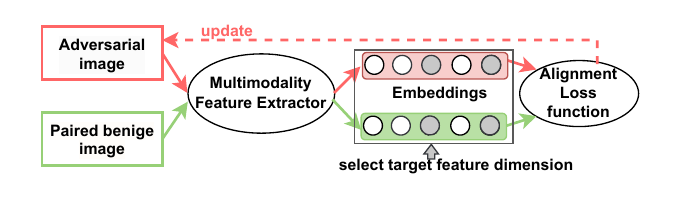}
    \caption{Alignment training to de-toxicity of the adversarial sample, with only selected target feature dimensions (in gray), i.e., \texttt{ImgD}, \texttt{TextD} and \texttt{CrossD}, involved in the alignment.}
    \label{fig:ad_overview}
\end{figure}

The adversarial sample is a benign-appearing image, e.g., a scenery image, but injected with harmful semantic information, such as the phrase \textit{``I want to make a bomb''}. One defense optimization strategy involves minimizing the distance between the embeddings of adversarial sample $\mathbf{F}_
{adv}$ and a benign sample $\mathbf{F}_{ben}$, and accordingly update the adversarial sample (in Figure~\ref{fig:ad_overview}). The paired benign image is injected with the friendly text, e.g.,  \textit{``peace and love''}. To study the effects of our identified modality features, we only select the target feature index $I$ for alignment training, i.e., \texttt{ImgD}, \texttt{TextD}, and \texttt{CrossD}. The alignment loss is $\mathcal{L} = \|\mathbf{F}_{adv}[:, I] - \mathbf{F}_{ben}[:, I]\|_2$. Finally, the optimized adversarial sample is then adopted to attack a VLM, LLaVA-1.5-7b~\citep{liu2023improvedllava}. We use the LLM-as-a-Judge to evaluate the generated response from the VLM, where DeepSeek-V3~\citep{deepseekai2024deepseekv3technicalreport} is required to generate a binary label indicating whether the attack is successful. Supposedly, the features containing more information related to the malicious semantics will contribute most to the attack defense.


\begin{table}[h]
    \centering
    \resizebox{0.45\textwidth}{!}{%
    \begin{tabular}{r|ccc}
    \toprule
    \textbf{Target feature} &\texttt{ImgD} & \texttt{TextD} & \texttt{CrossD} \\
    \midrule
    \textbf{Success Rate} ($\downarrow$) & 62.71\% & 24.89\% & 35.44\% \\ 
    \bottomrule
    \end{tabular}
    }
     \caption{Success rate for adversarial attacks with different target features involved in the alignment training. The success rate for original adversarial samples without alignment training is 73.26\%, while for randomly selected features is 54.28\%.}
     \label{tab:adversarial}
\end{table}
\begin{figure*}[ht]
    \centering 
    \includegraphics[width=0.85\textwidth,trim={50 0 50 0},clip]{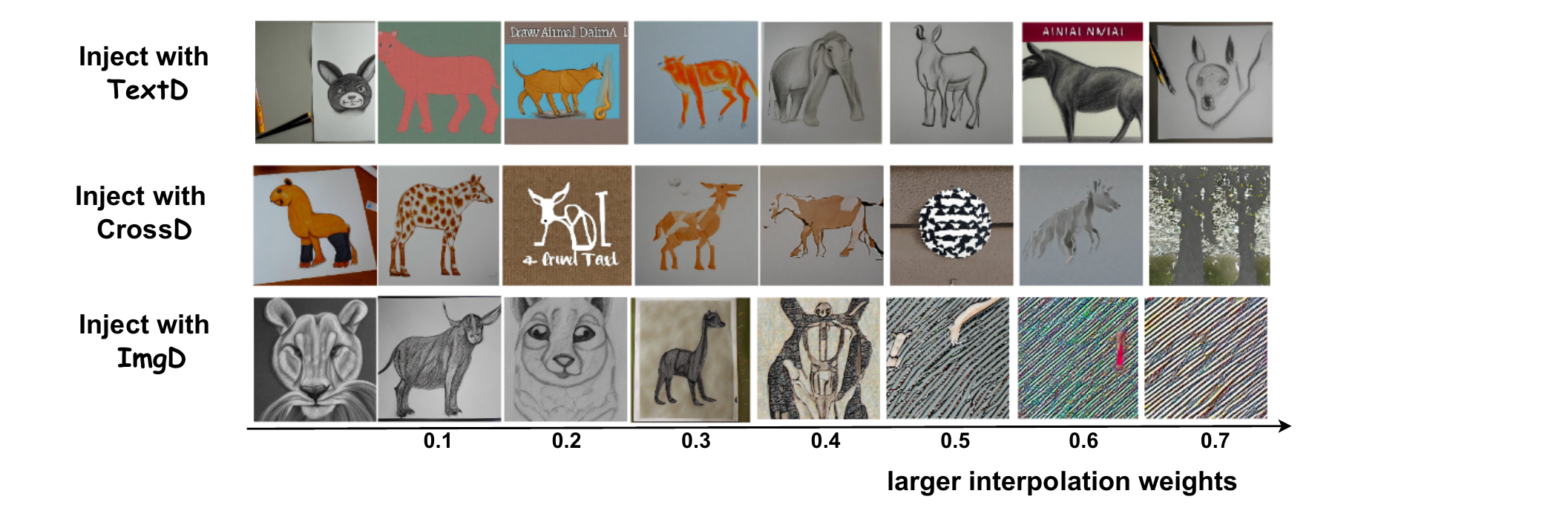}
    \vspace{-5mm}
    \caption{Generated new images from the VLM with the text prompt \textit{``Please draw an animal"} and varying levels of intervention from a reference image (horse). From left to right, the interpolation weights $\alpha$ range from 0.0 to 0.7. Images generated with \texttt{TextD} typically depict clear main subjects (horse) without transferring the visual background details from the reference image. In contrast, injection of \texttt{ImgD} introduces low-level visual details as well as image distortions when $\alpha$ is large.}
    \label{fig:mm_gen}
\end{figure*}
\begin{figure}[htbp]
    \centering 
    \includegraphics[width=0.35\textwidth,trim={40 50 20 20},clip]{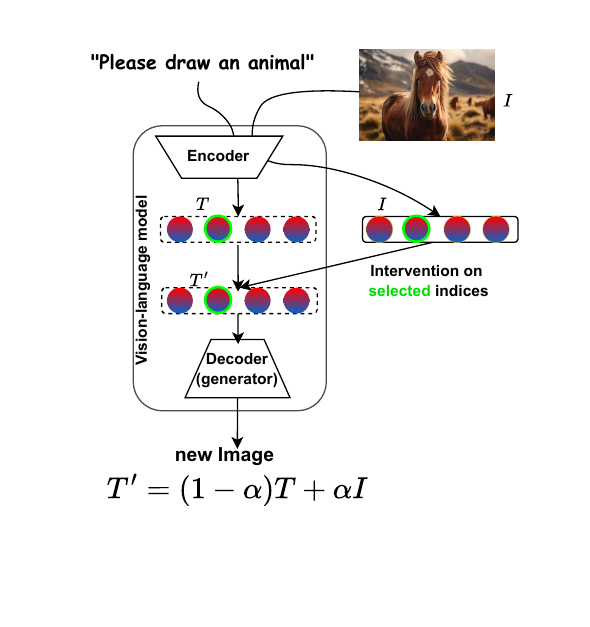}
    \vspace{-5mm}
    \caption{The reference image $R$ is used for modality-specific control over text-to-image generation process.}
    \label{fig:t2i_overview}
\end{figure}
\noindent\textbf{Results.} The attack success rates are shown in Table~\ref{tab:adversarial}. We select the same number from \texttt{ImgD}, \texttt{TextD}, \texttt{CrossD} to be involved in alignment training, as well as randomly sample the same number of features across the three feature sets as a baseline.  We attack the VLM repeatably for 100 times per sample, and we have generated 50 adversarial samples.  We observe that (i) by comparing with the success rate of original adversarial samples, the alignment training with any selected features defense the attacks to some extent; (ii) using \texttt{TextD} yields the best defense performance, followed by \texttt{CrossD} and \texttt{ImgD}. This can be explained by the fact that the adversarial information primarily stems from undesirable textual semantics. And \textbf{it demonstrates that \texttt{TextD} effectively captures most of the semantic content.} In contrast, \texttt{CrossD} captures partial semantics, while \texttt{ImgD} is the least related to semantic information, resulting in minimal benefits for such modality-specific jailbreak defense.

\subsection{Controllable Text-to-Image Generation}
\label{subsec:mul_gen}
Despite the impressive capabilities of text-to-image generation models~\citep{yu2024spae,koh2024generating,swamy2024multimodn}, their internal mechanisms for bridging linguistic semantics and visual details remain poorly understood. A key challenge is disentangling how modality-specific features influence the fidelity and controllability of generation. Therefore, we conduct a feature intervention experiment during the generation of Stable Diffusion v2~\citep{Rombach_2022_CVPR}. \\

\noindent\textbf{Intervention.} 
The process is depicted in Figure~\ref{fig:t2i_overview}. We investigate the generation process by intervening in different modality-specific features in Stable-Diffusion-v2~\citep{Rombach_2022_CVPR}, i.e., the shown VLM with an encoder and decoder (generator). 
The input text prompt is \textit{“Please draw an animal”}. The encoder generates an embedding $\mathbf{T}$, representing the original multimodal embedding ready for generation. Additionally, we provide a reference image; here is a horse - processed through the same encoder, producing a reference embedding $\mathbf{R}$. To control the generation through modality-specific feature intervention, we interpolate $\mathbf{T}$ only at the specified indices defined by MDS. The final multimodal embedding is computed as:
$\mathbf{T'}[I] = \alpha \mathbf{T}[I] + (1 - \alpha) \mathbf{R}[I]$, where operations are applied exclusively to the feature indices defined by $I$, i.e., \texttt{TextD}, \texttt{CrossD}, and \texttt{ImgD}.\\
\noindent\textbf{Results.} We feed $\bm{T'}$ to the generator of the VLM with different $\alpha$ ranging from 0 to 0.7 with an interval of 0.1. The generated images with the selected indices correspond to \texttt{TextD}, \texttt{CrossD}, and \texttt{ImgD} are shown in Figure~\ref{fig:mm_gen}. The results clearly demonstrate that larger interventions on \texttt{TextD} lead to stronger control over high-level semantic concepts—for example, the generated image more distinctly resembles a horse (head). All these generated images injected by \texttt{TextD} typically depict clear main subjects without transferring visual background details from the reference image. In contrast, interventions on \texttt{ImgD} result in more visual details from the reference image being preserved, such as non-white and fur-like patterned background, which are visible in \texttt{ImgD} when $\alpha \geq 0.3$. To better contrast the effects of \texttt{ImgD} and \texttt{TextD}, we also use a reference image with a horse as the main subject, but in different styles/backgrounds. More results are shown in Figure~\ref{fig:t2i_ref2}.

\section{Conclusion}
\label{sec:conclusion}
In this study, we explored the monosemanticity of features within VLMs to elucidate the commonalities and distinctions across visual and linguistic modalities. Specifically, we successfully categorized multimodal features according to their dominant modality. Our proposed embedding-based interpretability metrics fill the gap in multimodal monosemanticity assessment. Moreover, we designed lightweight probing and editing methods based on modality-specific features and demonstrated great potential in mitigating gender bias, defending against adversarial attacks, and enabling controllable multimodal generation.
\section*{Limitation}
While our work provides valuable insights into modality-specific feature analysis in vision-language models, several limitations warrant discussion.
First, we did not conduct human studies to validate our interpretability metrics. Although our embedding-based metrics align with existing interpretability tools, direct human evaluation could provide stronger evidence that our categorizations match human cognitive interpretations of modality dominance.
Second, our experiments focus exclusively on CLIP-family models. The generalizability of our findings to other vision-language architectures (e.g., BLIP or autoregressive VLMs) remains an open question. Different architectural designs may exhibit distinct modality gap characteristics that require adapted analysis methods.

\section*{Acknowledgment}
This work was supported in part by the UK Engineering and Physical Sciences Research Council through a Turing AI Fellowship (grant no. EP/V020579/1, EP/V020579/2) and the Prosperity
Partnership scheme (grant no. UKRI566). The authors also acknowledge the use of the King’s
Computational Research, Engineering, and Technology Environment (CREATE) at King’s College
London.

\bibliography{custom}
\newpage
\appendix
\clearpage
\newpage
\section{Appendix}
\subsection{Implementation for Monosemanticity Tools}
\label{app:extract_mono}
The three monosemantic tools, DeCLIP, Multimodal SAE, and Multimodal NCL are all on top of the canonical ViT-B-32 CLIP~\footnote{\url{https://github.com/openai/CLIP}} model from OpenAI~\citep{radford2021learning}, with ResNet50. The four methods (including CLIP) share the same model structures but are trained with different training objectives. We load them by feeding the checkpoints using the \texttt{open\_clip.create\_model\_and\_transforms} function in the published \url{https://github.com/mlfoundations/open_clip}.

The feature dimensions of the output features from the image encoder and text encoder are both 1024, the same for CLIP, DeCLIP, and Multimodal NCL. To retain the multimodal representation efficiency in downstream tasks, we have trained the SAE and NCL to reach a very small reconstruction loss for the original features $z$ from CLIP. The dataset for NCL and SAE training is the train split (around 2900k image-text pairs) from \texttt{cc3m-wds}\footnote{\url{https://huggingface.co/datasets/pixparse/cc3m-wds}}. We train the two variants, i.e., SAE and NCL, on top of the pretrained CLIP using a single 3090 GPU.

\paragraph{DeCLIP.}
We use the checkpoint released in \url{https://github.com/Sense-GVT/DeCLIP} to extract the last layer features, $z_i$ and $z_t$.

\paragraph{Multimodal SAE.}
We insert an SAE model to map the original feature into a sparse latent space, i.e., $z^{d}\rightarrow z^{n}$, with top-k latent as nonzero values. Empirically, we found that when $n=d$ and $k=32$, we can get the best results to balance the sparsity and downstream task performance. Such an SAE model (shared parameter) is inserted at the end of the image and text encoder in CLIP. 

\begin{figure}[h]
    \centering
    \includegraphics[width=0.95\linewidth]{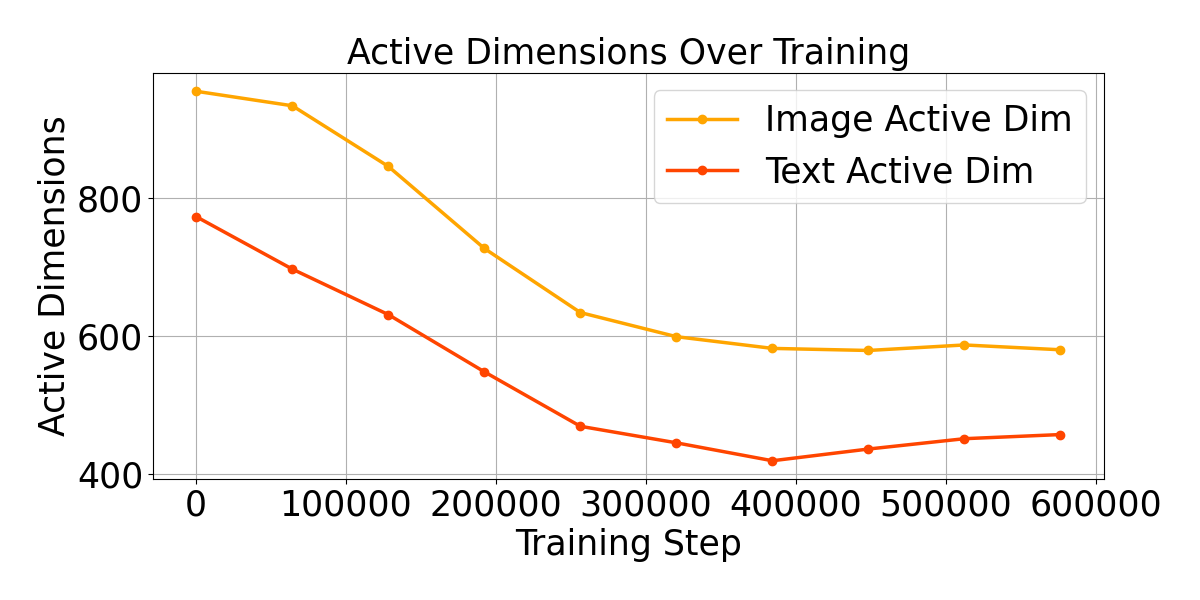}
    \caption{The changes of active dimensions over \textbf{SAE} training.}
    \label{fig:act_dims_sae}
\end{figure}

\begin{lstlisting}[language=Python, caption= , basicstyle=\footnotesize\ttfamily]
def get_sae_embedding(self, z):
    z = self.encoder(z)
    z_sae = F.relu(z)
    vals, ids = z_sae.topk(self.k, dim=1)
    z_sae = torch.zeros_like(z_sae)
    z_sae.scatter_(1, ids, vals)
    return z_sae
\end{lstlisting}
Inspired by~\cite{gao2024scaling}, we train the SAE until the sparsity (the inactive dimension)  of image features and text features doesn't increase (the same stop criteria for NCL). Noting that there are many zero values in $z^\text{sae}$, we remove those zero activity features (called dead latents in~\citep{gao2024scaling}) for further studies.  We show the changes of active dimensions of image features and text features in Figure~\ref{fig:act_dims_sae}.

\begin{figure}[h]
    \centering
    \includegraphics[width=0.98\linewidth]{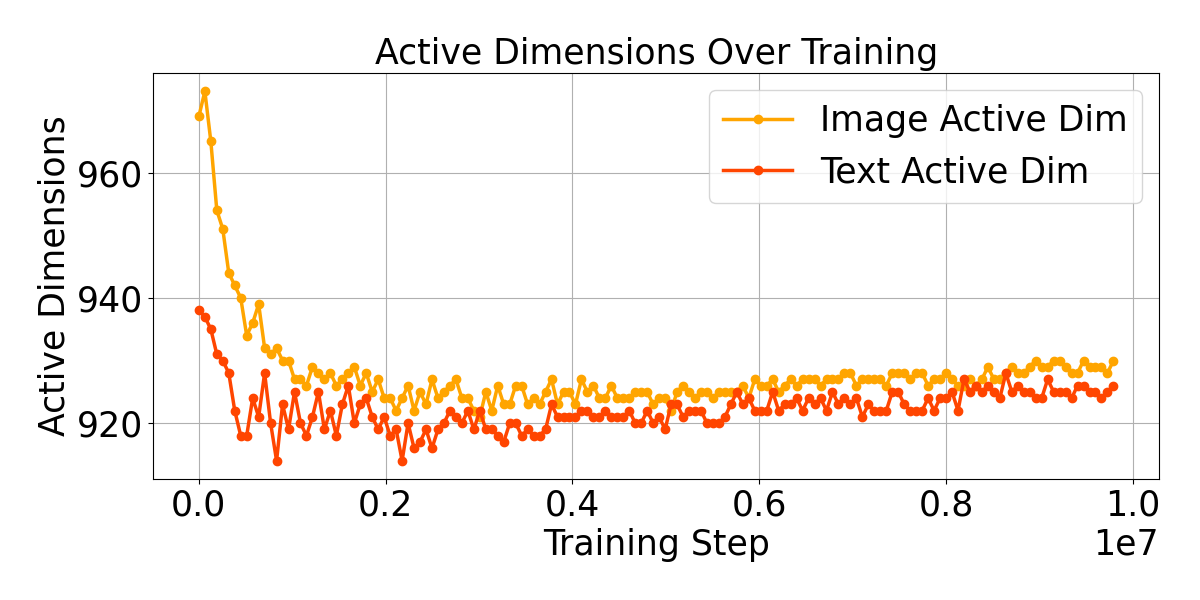}
    \caption{The changes of active dimensions over \textbf{NCL} training.}
    \label{fig:act_dims_ncl}
\end{figure}

\paragraph{Non-negative Contrastive Learning (NCL).}
We add the NCL block, i.e., the projector, after obtaining $z_i$ and $z_t$ from the image encoder and text encoder. The training loss is shown in Eq.~\ref{eq:ncl_loss}.
\begin{lstlisting}[language=Python, caption= , basicstyle=\footnotesize\ttfamily]
self.projector = nn.Sequential(
    nn.Linear(embed_dim, embed_dim),
    nn.LayerNorm(embed_dim),
    nn.ReLU(),
    nn.Linear(embed_dim, embed_dim),
)  
z_ncl = self.projector(z)
\end{lstlisting}
Similarly, the activated dimensions for image features and text features decrease and are then flattened (shown in Figure~\ref{fig:act_dims_ncl}.) By comparing with Figure~\ref{fig:act_dims_sae}, we noticed that the features in SAE is much more sparse than those in NCL.
\subsection{Implementation of MDS}
\label{app:mds}
Based on the trained CLIP, CLIP+SAE, CLIP+NCL, and DeCLIP, we feed the test split of \texttt{cc3m-wds} dataset to these pretrained models, around  15k image-text pairs to calculate MDS, according to Eq.(\ref{eq:mds}). The features are the last-layer output from the text and image encoder. We tried to calculate the normalization of $z_i$ and $z_t$, but found that it makes little difference to the final results. It could be attributed to the existing normalization technique in image and text encoders in CLIP. 

\subsection{Interpretability Evaluation}
\label{app:mono}
\subsubsection{Implementation}
\paragraph{Embedding models $h$ for activated image/text samples.} Our interpretability metrics, i.e., \textit{\textbf{EmbSim}} and \textbf{\textit{WinRate}} are based on the embeddings of active image/text samples by each feature. We need the embedding models to obtain these embeddings, i.e., $Z^{+}$ and $Z^{-}$. We use the Vision Transformer (\href{https://huggingface.co/google/vit-base-patch16-224-in21k}{ViT-B-16-224-in21k}) for image embeddings and the Sentence Transformer (\href{https://huggingface.co/sentence-transformers/all-MiniLM-L6-v2}{all-MiniLM-L6-v2}) for text embeddings. The goal here is to derive the general and effective image and text embeddings, so we can also use the image encoder and text encoder from CLIP. 

\subsubsection{Results}
\label{app:mono_results}
\subsection{Qualitative evaluation results}
\label{app:crossD_qualitative}
\begin{figure*}[h]
\centering
\includegraphics[width=0.65\linewidth]{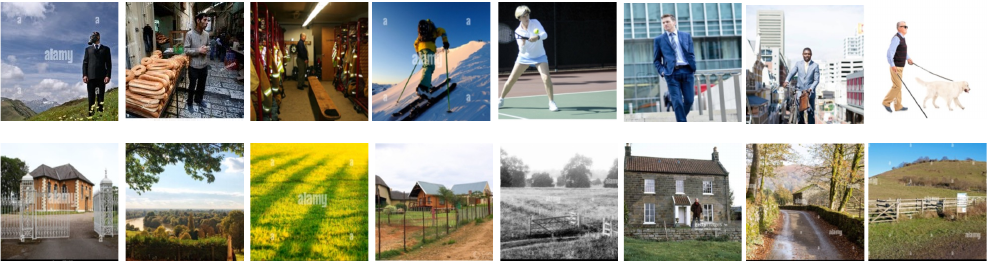}
\label{fig:act_img_crossm}
\vfill
\resizebox{0.65\textwidth}{!}{%
\begin{tabular}{p{6.6cm}|p{6.8cm}}
\toprule[1pt]
\multicolumn{1}{c|}{\textit{\textbf{Feature-6:  Actions performed by individuals}}}& \textit{\textbf{Feature-47: Outdoors Scenery}}\\
\midrule
\textcolor{blue}{Young man} working on invention in a warehouse.&A stile on a \textcolor{blue}{public footpath} overlooking the village on a frosty autumn morning. \\
\midrule
\textcolor{blue}{cricketers exercise} during a practice session.&A private chapel, and the wrought iron gates in the \textcolor{blue}{grounds}. \\
\midrule
\textcolor{blue}{Cricket player} checks his bat during a training session. & Train track: a man blending in with the \textcolor{blue}{scenery} as he stands on a railway track near a river 
\\
\midrule
\textcolor{blue}{Basketball coach} watches an offensive possession from the sideline during the second half. &surveying the \textcolor{blue}{scene}: people look out over loch today on a warm day in the village \\
\bottomrule
\end{tabular}
}
\vspace{-3mm}
\captionof{figure}{\footnotesize Activated images and texts by \texttt{CrossD} features. Top image row (feature 6): activities performed by individuals. Bottom image row (feature 47): scenery outside the doors. Text in \textcolor{blue}{blue} aligns with visual concepts.}
\vspace{-3mm}
\label{tab:act_texts_crossm}
\end{figure*}

\textbf{\texttt{CrossD} (the majority features) capture shared semantics across modalities.} 
Different from modality-specific features, \texttt{TextD} and \texttt{ImgD}, 
\texttt{CrossD} features capture common concepts that could be expressed in both visual and language modalities. We randomly select two \texttt{CrossD} features and show their top activated images and texts. As shown in Figure~\ref{tab:act_texts_crossm},
Feature-6 mostly activates scenes involving individuals performing activities, especially outdoor activities, and feature-47 captures general outdoor environments. The coherence across both modalities reflects successful alignment, which is consistent with multimodal training objectives.

\paragraph{MDS for different methods}
\textbf{MDS with monosemanticity enhancements.} With the monosemanticity-improving models (SAE and NCL), we hypothesize that modality purity will become more pronounced, making dominant modality assignments more meaningful. To validate this, we calculate the MDS and visualize the distributions of the three feature groups across models in Figure~\ref{fig:mds}. Interestingly, we find that CLIP, which is only trained on an image-text contrastive learning objective, contains a spectrum of features with different modality dominance. 
Specifically, its distribution skews towards the image modality, and this trend is consistent across all models. 
DeCLIP, on the other hand, shows a more balanced and less centered distribution. This suggests that DeCLIP, through self-supervision, extracts more modality-specific features, which might be overlooked by pure vision-language contrastive models like CLIP. 
The extracted features on top of NCL and SAE also exhibit less skewness, with SAE showing the most balanced distribution, indicating its strong capability to extract diverse monosemantic features. 
\begin{figure*}[h]
  \centering
  \includegraphics[width=0.89\linewidth,trim={130 120 140 130},clip,]{MDS_v2.pdf}
  \caption{Modality Dominance Score (MDS) distributions of three feature categories for different VLMs.}
  \label{fig:mds}
\end{figure*}

\paragraph{EmbSimi and WinRate for Monosemanticity measurement.} Firstly, we show the complete results for \textit{\textbf{EmbSmi}} and \textit{\textbf{WinRate}} in the Table~\ref{tab:interpret_metrics_allmodels}. 
\begin{table}[h]
    \centering
 
    \resizebox{0.45\textwidth}{!}{%
    \begin{tabular}{lcc|ccc}
    \toprule
    \textbf{Models}& \multicolumn{2}{c|}{\textbf{\textit{EmbSim}}} & \multicolumn{2}{c}{\textbf{\textit{WinRate}}}\\
    \cmidrule{2-5}
    \textbf{Activated}$\rightarrow$ &Image & Text & Image & Text\\
    \midrule
    \rowcolor{gray!20}CLIP&  0.11&0.45& \underline{0.65}&0.59\\
    DeCLIP & 0.06&	-0.07& 0.61&	0.46\\
    CLIP+NCL & \underline{0.14}&	\underline{0.45}& \textbf{0.71}&\underline{0.60}\\
    CLIP+SAE & \textbf{0.17} &\textbf{0.74}&0.60&\textbf{0.61}\\
    \bottomrule
    \end{tabular}
    }
    \caption{ Average interpretability scores (by examining the top activated images/texts) for features extracted from VLMs.}
    \label{tab:interpret_metrics_allmodels}
\end{table}

\begin{figure}
    \centering
    \includegraphics[width=0.9\linewidth]{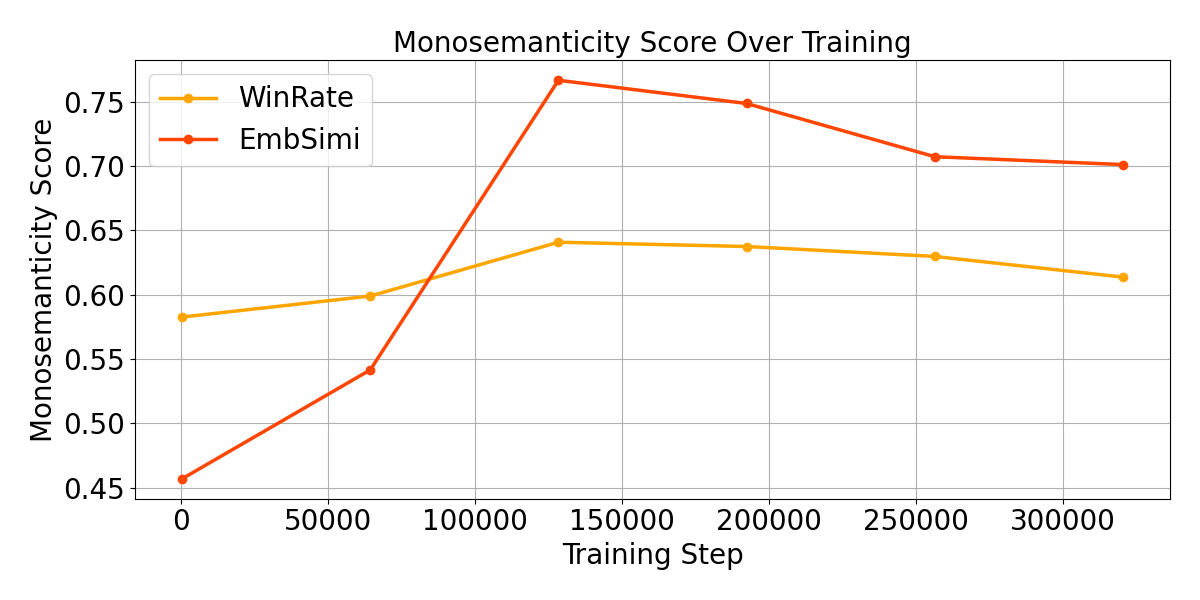}
    \includegraphics[width=0.9\linewidth]{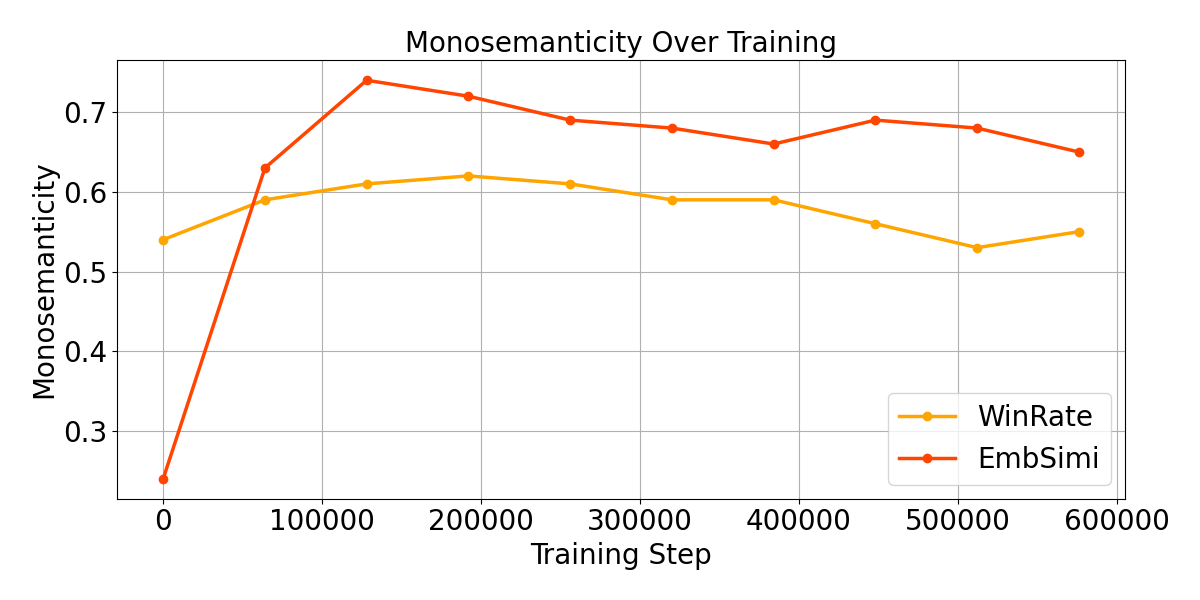}
    \caption{Monosemanticity (EmbSimi and WinRate) changes as training goes on. Upper is for CLIP+NCL, bottom is for CLIP+SAE.}
    \label{fig:mono_changes}
\end{figure}

\paragraph{The results of monosemanticity changes as training goes on.} We show the results of monosemanticity score changes as training goes on for both NCL and SAE in Figure~\ref{fig:mono_changes}. 

\begin{table}[h]
\centering
\label{tab:mono}
\resizebox{0.49\textwidth}{!}{%
    \begin{tabular}{r|>{\columncolor{gray!20}}c|c|c|c}
    \toprule[1pt]
     \textbf{Models}    & \textbf{CLIP}&\textbf{DeCLIP}&\textbf{CLIP+NCL}&\textbf{CLIP+SAE} \\
     \midrule
     \multicolumn{5}{c}{\textit{\textbf{Mono}} is \textit{\textbf{EmbSim}}}\\
     \midrule
    \textbf{Visual Mono} & -0.007 & \underline{0.009} & \textbf{0.043}& 0.005 \\
    \textbf{Textual Mono} & -0.017 & -0.001 &\textbf{0.210} &\underline{0.146}\\
    \midrule
    \multicolumn{5}{c}{\textit{\textbf{Mono}} is \textit{\textbf{WinRate}}}\\
    \midrule 
    \textbf{Visual Mono} &  -0.007&0.005&\underline{0.002}&\textbf{0.030}\\
    \textbf{Textual Mono} & -0.069&-0.059&\textbf{0.018}&\underline{0.016}\\
    \bottomrule[1pt]
    \end{tabular}
    }
    \caption{The visual and textual monosemanticity. A higher value indicates that \texttt{ImgD} captures more visual than linguistic features, and vice versa for \texttt{TextD}.}
\end{table}




\subsection{Implementations and More Results for Case Studies}
\label{app:case_study}
We provide the implementation details and more experimental results for the three case studies in the following.
\subsubsection{Case study 1: Understanding Gender Pattern in Different Modalities}
\paragraph{Datasets.} We select male and female images using a gender classifier \href{https://huggingface.co/touchtech/fashion-images-gender-age-vit-large-patch16-224-in21k-v3}{touchtech/fashion-images-gender-age-vit-large-patch16-224-in21k-v3} from cc3m-wds validation set.  We have both input images and text; the original gender classification accuracy is 83.4\% and 73.4\%, respectively. 
\paragraph{Classification.} As the intervened features are not compatible with existing pretrained text or text classifiers, we compare these features with the golden feature from male and female data. Specifically, we randomly select a female/male image with classification logits larger than 0.9 (ensuring the gender patterns are obvious) as the reference features. We use the same embedding models in \S\ref{app:mono}, i.e., Vision Transformer and Sentence Transformer as the encoder and encode both intervened feature and golden feature. The intervened feature is labeled with the same label as the reference image, for which its distance in encoder space is smaller.

\paragraph{Intervention.} There are different number of \texttt{ImgD} and \texttt{TextD} for a given representation of input sample. To avoid the effects of different numbers of removal features, we remove (set the corresponding dimension as zero) the minimal number between \texttt{ImgD} and \texttt{TextD}, and remove the same number of randomly selected features as a baseline. 

\paragraph{\texttt{TextD} in male concepts.}
We also cluster different male descriptions according to the percentage of \texttt{TextD} features among all their top-20 activated features, and we calculate the frequency of the top7 tokens in each cluster shown in Table~\ref{tab:male_sents_cluster}. We remove the gendered personal pronouns, e.g., he, she, woman, man, boy, girl, and only focus on how gender-neutral concepts represent the gender.  With more \texttt{TextD} injection, the textual descriptions become more sports-related, such as coach, basketball, soccer; while the sentences with less activated \texttt{TextD} have top words, such as party, hip, game, smile, home. This trend is consistent with the social stereotype that males are more active in sports activities.
\begin{table*}[h]
    \centering
    \resizebox{0.8\textwidth}{!}{%
    \begin{tabular}{c|c}
\toprule
   \textbf{Percentage of} \texttt{TextD}   & \textbf{Top8 words in male-related textual description}\\
\midrule
   0.1      & attends, party, hip, game, comedian, city, black, artist \\
   0.12 & smile, made, blue, outside, looks, home, got, book \\
   0.18 & artist, player, film, pop, performs, festival, young, suit\\
   0.24 & player, football, basketball, team, game, portrait, holding, gym\\
\bottomrule
    \end{tabular}
    }
        \caption{\footnotesize Representative words in male-related descriptions with different percentages of \texttt{TextD}.}
    \label{tab:male_sents_cluster}
\end{table*}

\subsubsection{Case Study 2: Defending Modality-Specific Adversarial Attacks}
\paragraph{Models}
We employed the same ViT-B-32 CLIP as in \S\ref{app:extract_mono} as the multimodality feature extractor shown in the Figure~\ref{fig:ad_overview} to extract 1024-dimension features, so we use the categorized \texttt{TextD}, \texttt{ImgD} and \texttt{CrossD} calculated before.  We use LLaVA-1.5-7b as the attacked VLM~\citep{liu2023improvedllava}. The whole process of defensing adversarial attacks is two steps:
\begin{itemize}
    \item \textbf{Generating adversarial images by injecting harmful requests.} We have a benign scenery image and a list of 50 harmful requests. Firstly, we create an image with a white background with the text saying the one piece of harmful request, as the contrast image. Then, we apply the alignment training by minimizing the distance between the benign image and the contrast image in the embedding space of the image encoder. The benign image is thus being injected with harmful semantics, denoted as $\mathbf{F}_{adv}$. 
    \item \textbf{Defending the adversarial attacks.} To remove the toxicity of the adversarial samples, we employ the alignment training shown in Figure~\ref{fig:ad_overview} by updating the embeddings of the adversarial samples. Specifically, we only select the target features, i.e., the \texttt{ImgD}, \texttt{TextD}, and \texttt{CrossD} to be involved in the training. 
\end{itemize}

When attacking the VLM, we feed the adversarial images/samples along with the text prompt, i.e., the corresponding harmful request injected into the adversarial sample. For each adversarial sample, we repeat the attack process 100 times. For comparison, we apply the original generated 50 adversarial samples to attack VLM, and the average success rate is 73.26\%; and the success rate of the (benign image - harmful request) is 10.00\%. We conducted five independent runs for each experiment to ensure statistical reliability. Results in the tables show mean values across runs, with relative standard deviations below 3\% for accuracy metrics.

\textbf{Computing resources cost} 
The experiments were conducted with a GPU with 48GB of memory. Adversarial sample generation requires approximately 4 GPU hours, while adversarial sample detoxification takes approximately 6 GPU hours.

\subsubsection{Case Study 3: Modality-Aware Control for Text-to-Image Generation}
\paragraph{Models.} We select Stable-Diffusion-v2 (\url{https://huggingface.co/stabilityai/stable-diffusion-2}) as our text-2-image generation model. As its image encoder (CLIP-ViT-H-14-laion2B-s32B-b79K) is not the same CLIP we used before, we recalculate the MDS distribution to derive the three categories of features. 


\textbf{More results.}
We present additional images generated by modifying the original multimodal representation through feature injection from a reference image. To emphasize the distinction between \texttt{ImgD} and \texttt{TextD}, we use two reference images of horses in different backgrounds and artistic styles. Specifically, we compare two sets of images where features from sketch and oil painting styles are injected using \texttt{ImgD}. We observe that images influenced by sketches tend to be predominantly black and white, while those influenced by oil paintings appear more colorful. In contrast, the images generated using \texttt{TextD} remain visually similar across both the sketch and oil painting settings.

\begin{figure*}[ht]
    \centering     \includegraphics[width=0.99\textwidth]{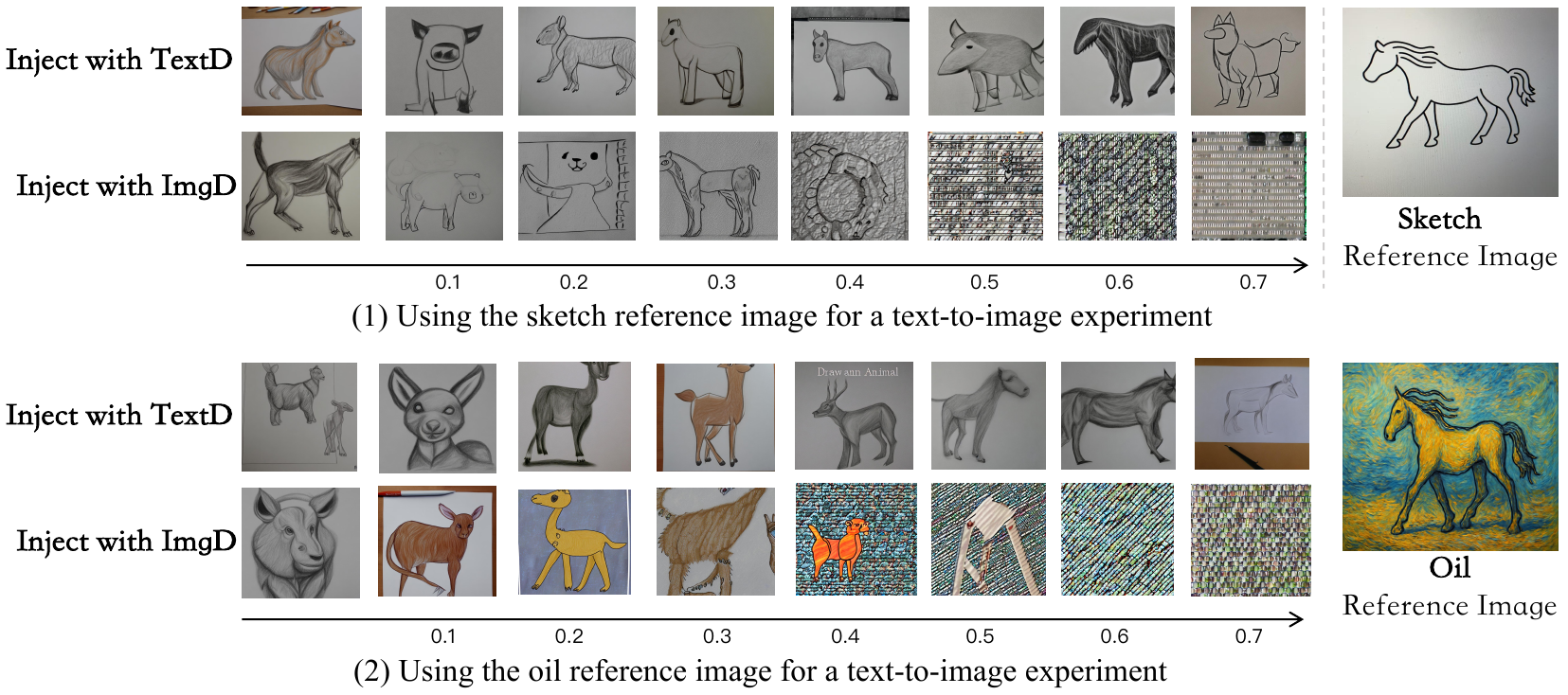}
    \caption{Generated new images from the VLM with the text prompt \textit{"Please draw an animal"} and varying levels of intervention from different reference images. We found that \textit{TextD} captures significant semantic information, such as shape, etc. 
    Notably, when a sketch is selected as the reference image, both \textit{imgD} and \textit{TextD} display sketch-like stylistic features. When oil-painting is chosen as the reference image, both \textit{imgD} and \textit{TextD} exhibit styles that resemble oil paintings. Comparatively, the stylistic differences between \textit{imgD} in conditions (1) and (2) are distinct: \textit{imgD} in (1) lacks color, whereas \textit{imgD} in (2) presents diverse coloration. Similar to Figure~\ref{fig:mm_gen}, \textit{TextD} does not affect low-level visual features, while \textit{ImgD} shows significant distortion at higher $\alpha$ values.}    
    \label{fig:t2i_ref2}
\end{figure*}

\end{document}